\documentclass[journal]{IEEEtran}
\usepackage{amsmath,bm}
\usepackage[pdftex]{graphicx}
\usepackage{comment}

\ifCLASSINFOpdf
\else
\fi
\begin{document}
%
\title{Weighted Sigmoid Gate Unit for \\ an Activation Function of Deep Neural Network}
%
%
%

\author{Masayuki~Tanaka,~\IEEEmembership{Member,~IEEE,}
\thanks{M. Tanaka was with 
the Artificial Intelligence Center, National Institute of Advanced Industrial Science and Technology, Japan.}
\thanks{Manuscript received April YY, 20XX; revised August YY, 20XX.}}

\maketitle

\begin{abstract}
 An activation function has crucial role in a deep neural network.
 A simple rectified linear unit (ReLU) are widely used for the activation function.
 In this paper, a weighted sigmoid gate unit (WiG) is proposed as the activation function.
 The proposed WiG consists of a multiplication of inputs and the weighted sigmoid gate.
 It is shown that the WiG includes the ReLU and same activation functions as a special case.
 Many activation functions have been proposed to overcome the performance of the ReLU.
 In the literature, the performance is mainly evaluated with an object recognition task.
 The proposed WiG is evaluated with the object recognition task and the image restoration task.
 Then, the expeirmental comparisons demonstrate the proposed WiG overcomes the existing activation functions including the ReLU.
\end{abstract}


%
\IEEEpeerreviewmaketitle

\section{Introduction}
In a deep neural network, an activation function plays critical roles.
A sigmoid function defined by $\sigma(x)=(1+e^{-x})^{-1}$ was classically used as the activation function. 
Currently, the Rectified Linear Unit (ReLU)~\cite{nair2010rectified} defined by $f(x)={\rm max}(x,0)$ is widely used for the activation function.
Many activation functions have been proposed to replace the ReLU in the literatures~\cite{Swish2017, Sil2017, Softplus2010, klambauer2017self, maas2013rectifier, he2015delving, ELU2016, ThresholdedReLU2015, van2016conditional}.

In those activation functions, a sigmoid gating approach is considered as promising approach.
The Sigmoid-weighted Linear Unit (SiL)~\cite{Sil2017} defined by $f(x) = x \cdot \sigma(x)$ has been proposed for reinforcement learning as an approximation of the ReLU.
So-called Swish unit~\cite{Swish2017} defined by $f(x)=x \cdot \sigma(\beta x)$, where $\beta$ is a trainable parameter, has been proposed.
The SiL can be considered as a special case of Swish with $\beta = 1$. 
In addition, the Swish become like the ReLU function for $\beta \rightarrow \infty$. 
They have shown the Swish unit outperforms other activation functions in object recognition tasks. 
It is known that Sigmoid gating is used in powerful tool of a long short-term memory (LSTM)~\cite{hochreiter1997long}.

Researchers mainly focused the activation function for a scalar input.
For the vector input, the scalar activation function is element-wisely applied.
Recently, sigmoid gaiting approaches for the vector inputs have been proposed~\cite{van2016conditional,wu2016multiplicative,dauphin2016language}.

In this paper, we propose a sigmoid gating approach as a activation function for vector input.
The proposed activation unit is called a weighted sigmoid gate unit (WiG). 
The proposed WiG consists of the multiplication of the inputs and weighted sigmoid gate as shown in Fig.~\ref{fig:WiGunit}.
The SiL~\cite{Sil2017} and the Swish~\cite{Swish2017} can be considered as a special case of the proposed WiG.

In the literature like~\cite{Swish2017}, previously activation functions have been only evaluated with an object recognition task. 
In this paper, we evaluate the proposed WiG with the object recognition task and the image restoration task. 
Experiments demonstrate that the proposed WiG outperforms existing other activation functions in both the object recognition task and the image restoration task.

\section{Weighted Sigmoid Gate Unit (WiG)}

\begin{figure}[t]
    \centering
    \includegraphics[width=6cm]{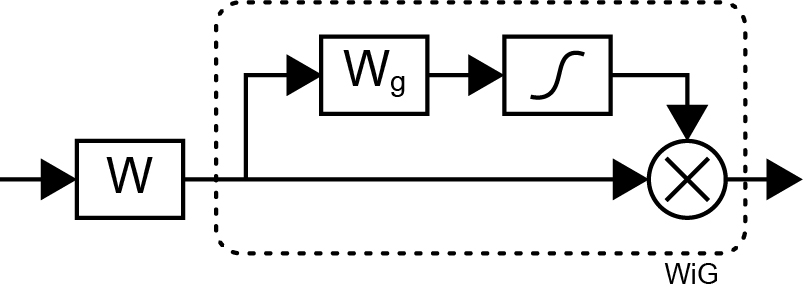}
    \caption{Proposed WiG, where the bias component is omitted.}
    \label{fig:WiGunit}
    
    \vspace*{3mm}
    
    \includegraphics[width=6cm]{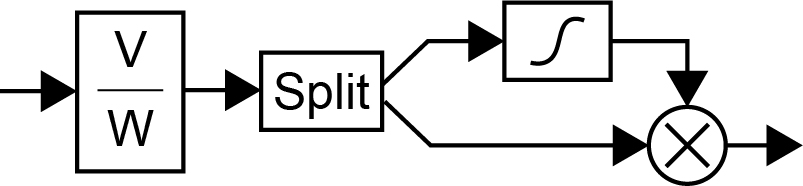}
    \caption{Combination of weighting unit and proposed WiG, where the bias component is omitted.}
    \label{fig:WWiGunit}
\end{figure}

A proposed weighted sigmoid gate unit (WiG) is expressed by element-wise product of the input and the sigmoid activation as
\begin{eqnarray}
 {\bm f}({\bm x}) &=& {\bm x} \odot \sigma( {\bm W}_g {\bm x} + {\bm b}_g ) \,,
 \label{eq:def_WiG}
\end{eqnarray}
 where ${\bm x} \in {\cal R}^N$ is $N$-dimensional input vector, 
 ${\bm W}_g \in {\cal R}^{N \times N}$ is $N \times N$ square weighting matrix, 
 ${\bm b}_g \in {\cal R}^N$ is $N$-dimensional bias vector, 
 $\sigma(\cdot)$ represents the element-wise sigmoid function,
 and
 $\odot$ represents element-wise products. 
Figure~\ref{fig:WiGunit} show a block diagram of the proposed WiG, where the bias component is omitted for the simplification.
The activation function usually follow the weighting unit. 
The WiG can be used as the activation function.

 In the deep neural network, the activation function usually follow the weighting unit.
 The combination of the weighting unit and the WiG shown in Fig.~\ref{fig:WiGunit} can be expressed as
 \begin{eqnarray}
  {\bm f}({\bm W} {\bm x}) &=& {\bm W}{\bm x} \odot \sigma( {\bm W}{\bm W}_g {\bm x} )
  \nonumber \\
  &=& {\bm W}{\bm x} \odot \sigma( {\bm V} {\bm x} )
  \,, \label{eq:def_WWiG}
 \end{eqnarray}
 where ${\bm W}$ is the weighting matrix, 
 ${\bm V} = {\bm W}{\bm W}_g$,
 and the bias components are omitted for the simplification.
 Equation~\ref{eq:def_WWiG} can be implemented as shown in Fig.~\ref{fig:WWiGunit}.
 This network is similar to that in~\cite{van2016conditional}.
 In terms of the calculation complexity, the networks of Figs.~\ref{fig:WiGunit} and \ref{fig:WWiGunit} are same.
 However, if we consider the parallel computation, the network of Fig.~\ref{fig:WWiGunit} is computationally efficient.
 For the training, the network of Fig.~\ref{fig:WiGunit} might be better, because the statistical properties of matrices ${\bm V}$ and ${\bm W}$ are expected to be different.

  
 The derivative of the WiG in~Eq.~\ref{eq:def_WiG} can be expressed by
 \begin{eqnarray}
  \frac{\partial {\bm f}}{\partial {\bm x}} &=&
  {\bm f} ({\bm W}_g {\bm x} + {\bm b}_g) \nonumber \\
  & &+ {\bm W}_g \cdot \{ {\bm x} \odot (1-\sigma( {\bm W}_g {\bm x} + {\bm b}_g )) \odot \sigma( {\bm W}_g {\bm x} + {\bm b}_g )\} \,. \nonumber \\
  \label{eq:def_WiG_der}
 \end{eqnarray}

 For the convolutional neural network, the proposed WiG can be expresses by
 \begin{eqnarray}
  {\bm F}({\bm X}) &=& {\bm X} \odot \sigma( {\bm w}_g * {\bm X} + {\bm B}_g ) \,,
  \label{eq:def_WiG_conv}
 \end{eqnarray}
  where 
  ${\bm X}$ is the feature map, 
  ${\bm w}_g$ is the convolutional kernel, 
  ${\bm B}_g$ is the bias map,
  and
  $*$ represents convolutional operator.
  

 \begin{figure}[t]
 \newlength{\miniwidth}
 \setlength{\miniwidth}{40mm}
 \newlength{\imgwidth}
 \setlength{\imgwidth}{.975\miniwidth}
 \newlength{\vsp}
 \setlength{\vsp}{1mm}
 
 \centering
 
 \hfill
 \begin{minipage}[t]{\miniwidth}
  \center
  \includegraphics[width=\imgwidth]{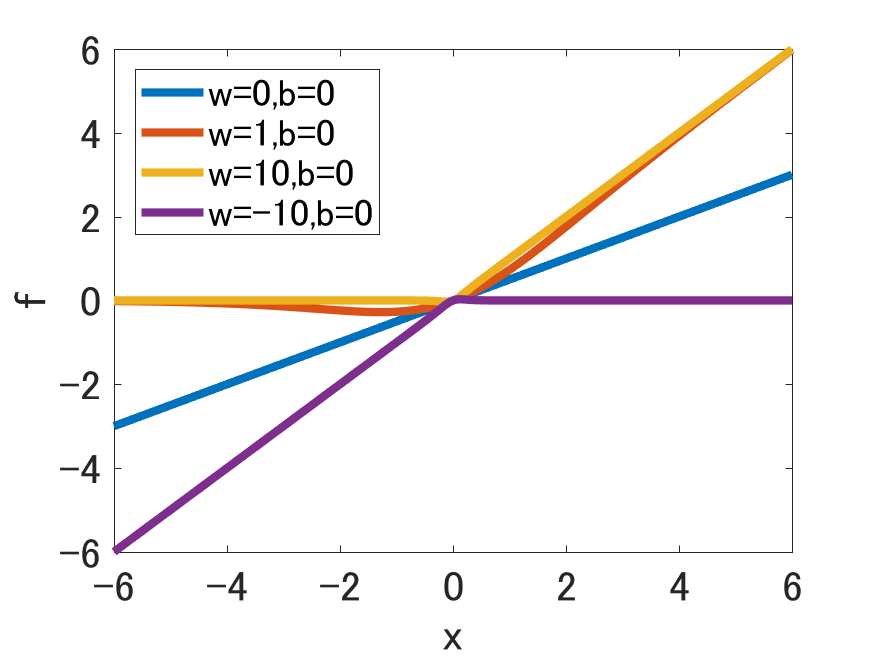}

  \vspace*{-\vsp}
  (a) Activation functions
 \end{minipage}
 \hfill
 \begin{minipage}[t]{\miniwidth}
  \center 
  \includegraphics[width=\imgwidth]{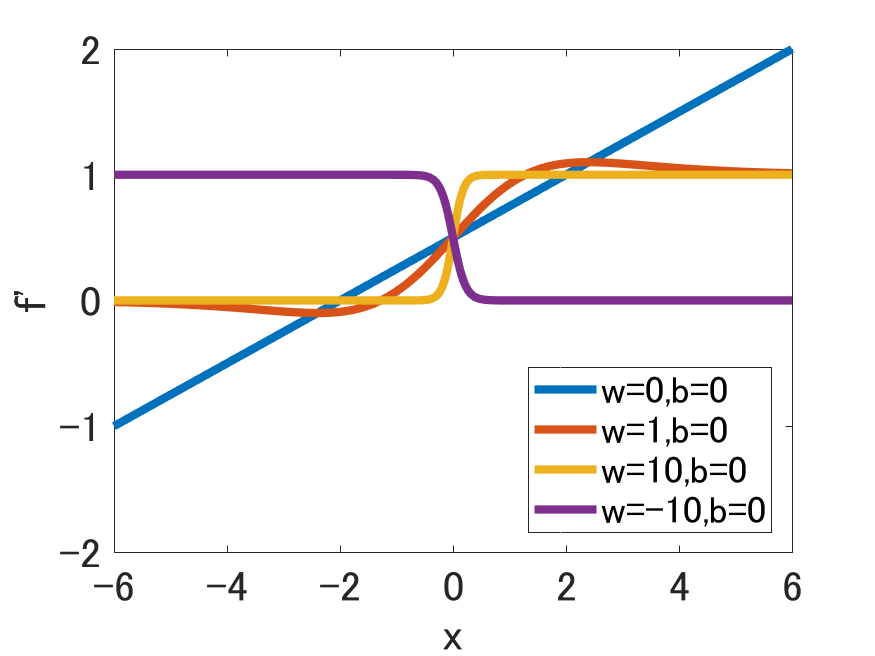} 

  \vspace*{-\vsp}
  (b) Derivatives
 \end{minipage}
 \hfill
 
 \caption{WiG functions and derivatives with $b=0$.}
 \label{fig:WiG_b0} 
 
 \vspace*{3.0mm}
  
 \hfill
 \begin{minipage}[t]{\miniwidth}
  \center
  \includegraphics[width=\imgwidth]{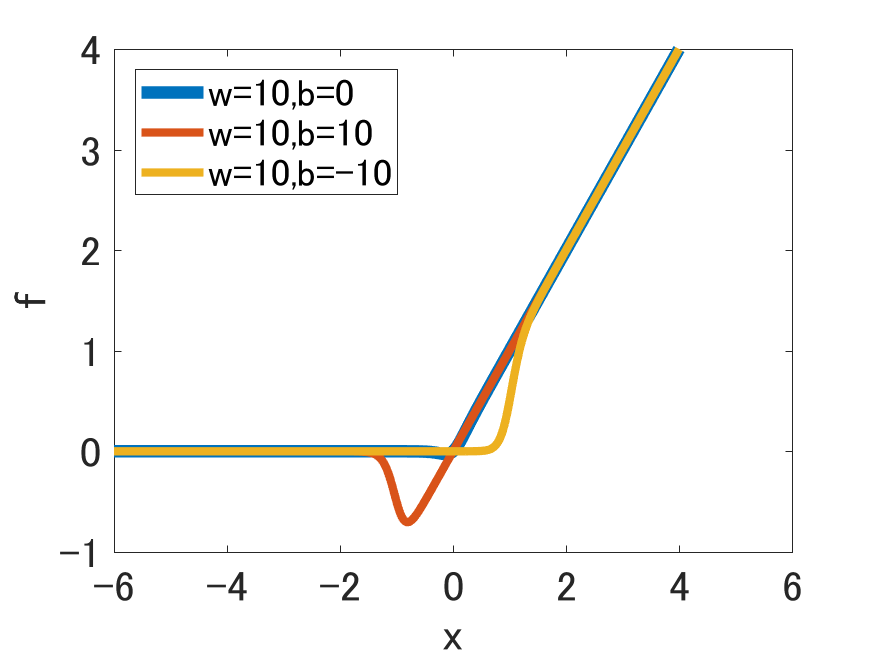}

  \vspace*{-\vsp}
  (a) Activation functions
 \end{minipage}
 \hfill
 \begin{minipage}[t]{\miniwidth}
  \center 
  \includegraphics[width=\imgwidth]{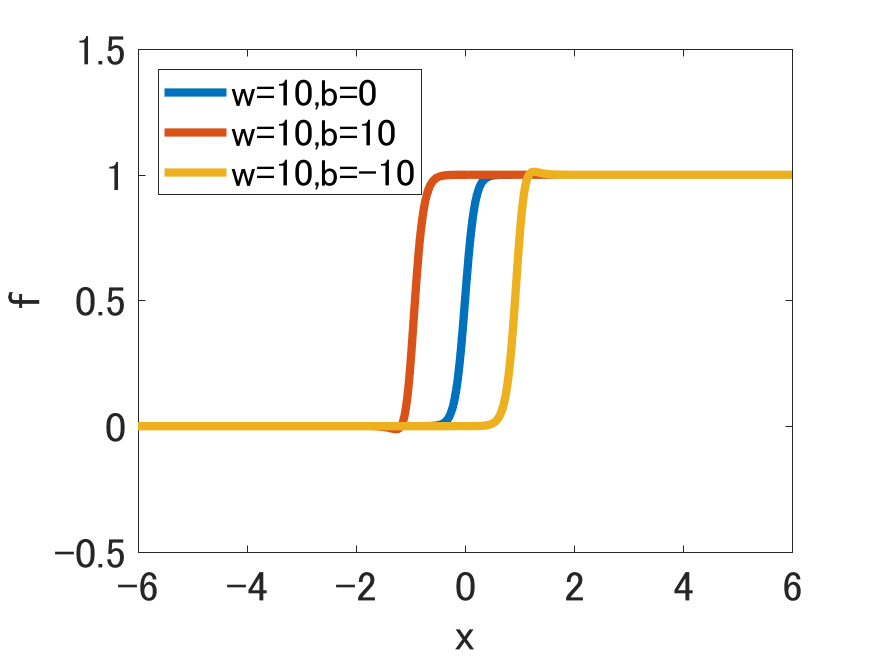} 

  \vspace*{-\vsp}
  (b) Derivatives
 \end{minipage}
 \hfill
 
 \caption{WiG functions and derivatives with $w=10$.}
 \label{fig:WiG_w10} 
\end{figure}

 \subsection{Simple Example}
 Here, we consider a simple scalar input case of WiG.
 Namely,
 \begin{eqnarray}
  f(x) &=& x \times \sigma (w x + b) \,.
  \label{eq:ssWiG}
 \end{eqnarray}　
 Figure~\ref{fig:WiG_b0} plots the graph of activation functions and derivatives of the WiG in Eq.~\ref{eq:ssWiG} for different gains of $w$, where the bias $b$ is set to zero.
 The SWISH~\cite{Swish2017} is a special case of the proposed WiG when the bias $b$ equals zero. 
 If the gain $w$ is 0, the proposed WiG is identical to the linear activation.
 As the gain $w$ is becoming positive large value, the proposed WiG function is becoming like the ReLU function.
 As the gain $w$ is becoming negative large value, the proposed WiG function is becoming like the negative-ReLU defined by ${\rm min}(x,0)$.
 It has been reported that the combination of the positive and the negative ReLUs is effective to improve the performance of the networks~\cite{kim2016onoffrelu,uchida2018coupledrelu}.
 
 Figure~\ref{fig:WiG_w10} plots the graph of activation functions and derivatives of the WiG in Eq.~\ref{eq:ssWiG} for different biases of $b$, where the gain $w$ is set to 10.
 Figure~\ref{fig:WiG_w10}-(b) clearly shows that the bias $b$ can controls the threshold of the derivatives.
 
 As shown in here, the proposed WiG activation function includes the ReLU~\cite{nair2010rectified} and the SWISH~\cite{Swish2017} as special cases.

\subsection{Initialization}
  The neural network model is highly non-linear model.
  Then, a parameter initialization is very important for optimization.
  The weighting parameters are usually initialized by zero-mean Gaussian distribution with small variance.
  The bias parameters are set to zero.
  
  However, we apply different initialization for the WiG parameters ${\bm W}_g$ and ${\bm b}_g$ in~Eq.~\ref{eq:def_WiG}.
  We initialize the scaled identical matrix $s {\bm I}$ for the weighting parameter ${\bm W}_g$ and zero for the bias parameter ${\bm b}_g$, so that the WiG is to be the Swish~\cite{Swish2017}, where $s$ is scale parameter and ${\bm I}$ is an identical matrix.
  If the scale parameter $s$ is large value, the WiG is initialized by the approximated ReLU. 
  This initialization with large scale parameter is very useful for the transfer learning from the network with widely used ReLU.
  When we train the network with the WiG from scratch, the scale parameter is set with one.
  Then, the WiG is initialized with the SiL~\cite{Sil2017}.

 \subsection{Sparseness constraint}
  A weight decay, which is regularization technique for the weight, is usually applied for the optimization.
  This technique can be considered as adding constraint $\lambda |{\bm W}|_2^2$ to the loss function, where $\lambda$ is constraint parameter, ${\bm W}$ is weighting parameter, and $|\cdot|_2$ represents $L_2$ norm.
  We introduce sparseness constraint for the WiG.
  The sparseness constraint of the WiG can be formulated by the $L_1$ norm of gate as
  \begin{eqnarray}
   L &=& \lambda_g | \sigma( {\bm W}_g {\bm x} + {\bm b}_g ) |_1 \,,
   \label{eq:SpConstraint}
  \end{eqnarray}
  where $\lambda_g$ is constraint parameter.
  The output of the sigmoid function can be considered as the mask.
  Eq.~\ref{eq:SpConstraint} is the sparseness constraint of the mask. 
  It also enforces the sparseness of the output of the WiG. 

\section{Experiments}
We experimentally demonstrate the effectiveness of the proposed WiG activation function comparing with existing activation functions: 
the ReLU~\cite{nair2010rectified},
the selu~\cite{klambauer2017self},
the ELU~\cite{ELU2016},
the softplus~\cite{Softplus2010},
the Leaky ReLU~\cite{maas2013rectifier},
the SiL~\cite{Sil2017},
the PReLU~\cite{he2015delving},
and
the Swish~\cite{Swish2017}.
Several activation functions are required parameters.
For those parameters, we used default parameters which are used in the original paper.
In related works~\cite{Swish2017,Sil2017}, they have only evaluated with object recognition task, while we evaluate the proposed WiG activation function with the object recognition task and the image restoration task\footnote{The reproduction code is available on http://www.ok.sc.e.titech.ac.jp/~mtanaka/proj/WiG/}.

\begin{figure}[t]
    \centering
    \includegraphics[width=8.5cm]{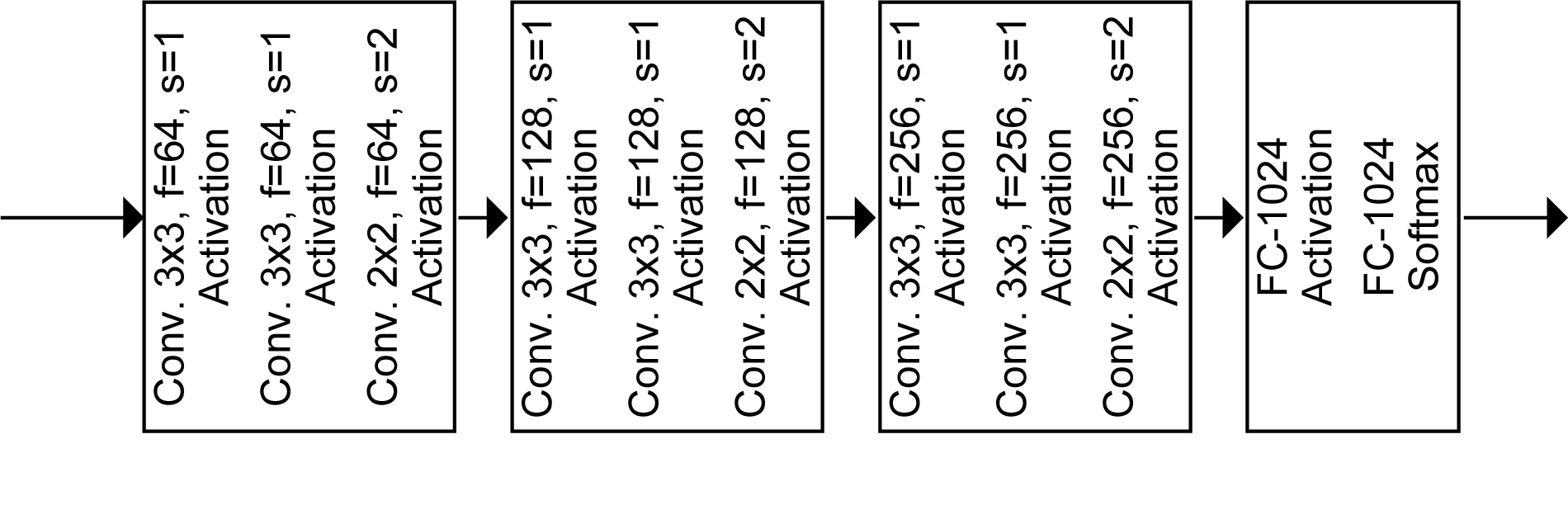}
    \caption{Object Recognition Network.}
    \label{fig:vgg}
\end{figure}

\begin{figure}[t]
    \centering
    \includegraphics[width=8.5cm]{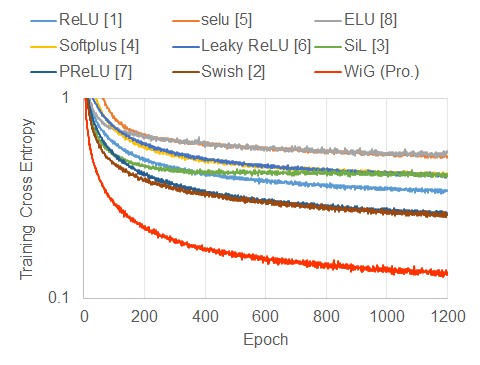}
    \caption{Training Cross Entropy of cifar10, where cross entropy is log-scale.}
    \label{fig:Cifar10Epoch}
\end{figure}

\begin{table}[b]
    \centering
    \caption{Comparison of the validation accuracy.}
    \label{tab:cifar}
    \begin{tabular}{c|cc} \hline
         & cifar10 & cifar100  \\ \hline
ReLU~\cite{nair2010rectified} & 0.927 & 0.653 \\
selu~\cite{klambauer2017self} & 0.899 & 0.572 \\
ELU~\cite{ELU2016} & 0.903 & 0.550 \\ 
softplus~\cite{Softplus2010} & 0.908 & 0.598 \\ 
Leaky ReLU~\cite{maas2013rectifier} & 0.918 & \underline{0.673} \\ 
SiL~\cite{Sil2017} & 0.919 & 0.638 \\ 
PReLU~\cite{he2015delving} & \underline{0.935} & \underline{0.678} \\ 
Swish~\cite{Swish2017} & \underline{0.935} & \underline{0.689} \\ 
WiG (Pro.) & {\bf \underline{0.949} } & {\bf \underline{0.742} } \\ \hline
    \end{tabular}
\end{table}

 \subsection{Object Recognition Task}
 Here, we evaluate the activation function with cifar data set~\cite{cifar}.
 For the object recognition task, we build a vgg-like network~\cite{VGG2015} as shown in~Fig.\ref{fig:vgg}, where we use a spatial dropout~\cite{tompson2015efficient} and a convolution pooling~\cite{springenberg2014striving} instead of a max pooling. 
 The dropout rates are increased for deep layer which is close to the output following paper~\cite{Ishii2016}.
 
 We built the networks with nine different activation functions as mentioned in the previous section. 
 Then, each network was trained with 1,200 epochs, where mini-batch size was 32 and we applied geometrical and photometric data argumentation.
 We used Adamax~\cite{Adam2015} for the optimization.
 The loss function is a categorical cross entropy.
 
 Figure~\ref{fig:Cifar10Epoch} shows a training cross entropy for each epoch.
 That comparison demonstrate that the network with the proposed WiG can rapidly learn and archive lower training cross entropy.
 The validation accuracy for cifar10 and cifar100 with different activation functions are summarized in Table~\ref{tab:cifar}, where the bold font represents the best performance and the underline represents higher accuracy than that of the ReLU. 
 First, the simple ReLU~\cite{nair2010rectified} has good performance.
 The proposed WiG demonstrates the best performance in both cifar10 and cifar100.
 We can find large improvement of the proposed WiG, esspecially for the cifar100 dataset.

 \subsection{Image Restoration Task}
 In this paper, we also compare the activation functions with an image restoration task.
 There are various image restoration tasks.
 The denoising task is known as important and essential image restoration task~\cite{liu2013single}.
 Therefore, we evaluate the activation function with image denoising task.

\begin{figure}[t]
    \centering
    \includegraphics[width=8.5cm]{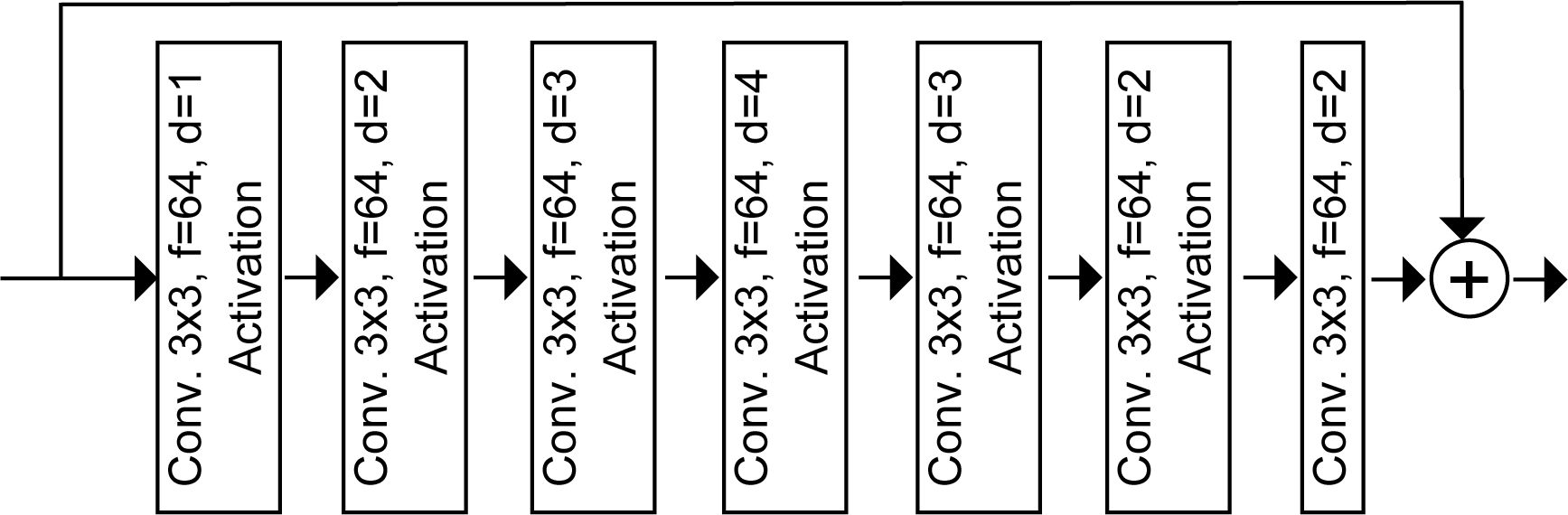}
    \caption{Denoising Network.}
    \label{fig:DenoisingNet}
\end{figure}

\begin{table}[b]
    \centering
    \caption{Comparison of the average PSNR of denoising results.}
    \label{tab:dconv_psnr}
    \tabcolsep=1.5mm
    \begin{tabular}{c|cccccc} \hline
Noise Level & 5 & 10 & 15 & 20 & 25 & 30  \\ \hline
ReLU~\cite{nair2010rectified} & 37.13 & 33.86 & 32.02 & 30.76 & 29.77 & 28.96 \\
selu~\cite{klambauer2017self} & 36.89 & 33.68 & 31.88 & 30.60 & 29.60 & 28.79 \\
ELU~\cite{ELU2016} & 37.07 & 33.63 & 31.86 & 30.60 & 29.65 & 28.86 \\ 
softplus~\cite{Softplus2010} & 36.41 & 33.42 & 31.67 & 30.45 & 29.50 & 28.69 \\ 
Leaky ReLU~\cite{maas2013rectifier} & 37.07 & 33.81 & 31.97 & 30.68 & 29.69 & 28.89 \\ 
Swish~\cite{Swish2017} & 37.13 & 33.70 & 31.91 & 30.69 & 29.72 & 28.93 \\ 
WiG (Pro.) & {\bf \underline{37.29} } & {\bf \underline{34.00} } & {\bf \underline{32.16} } & {\bf \underline{30.88} } & {\bf \underline{29.90} } & {\bf \underline{29.10} } \\ \hline
    \end{tabular}
\end{table}

\begin{table}[b]
    \centering
    \caption{Comparison of the average SSIM of denoising results.}
    \label{tab:dconv_ssim}
    \tabcolsep=1.1mm
    \begin{tabular}{c|cccccc} \hline
Noise Level & 5 & 10 & 15 & 20 & 25 & 30  \\ \hline
ReLU~\cite{nair2010rectified} & 0.9383 & 0.8971 & 0.8646 & 0.8378 & 0.8142 & 0.7932 \\
selu~\cite{klambauer2017self} & 0.9355 & 0.8929 & 0.8607 & 0.8328 & 0.8084 & 0.7855 \\
ELU~\cite{ELU2016} & 0.9362 & 0.8910 & 0.8601 & 0.8331 & 0.8094 & 0.7888 \\
softplus~\cite{Softplus2010} & 0.9337 & 0.8878 & 0.8547 & 0.8274 & 0.8041 & 0.7822 \\
Leaky ReLU~\cite{maas2013rectifier} & 0.9380 & 0.8959 & 0.8628 & 0.8350 & 0.8104 & 0.7888 \\
Swish~\cite{Swish2017} & 0.9377 & 0.8928 & 0.8621 &0.8370 & 0.8137 & 0.7928 \\
WiG (Pro.) & 
{\bf \underline{0.9390} } & 
{\bf \underline{0.8993} } & 
{\bf \underline{0.8679} } & 
{\bf \underline{0.8412} } & 
{\bf \underline{0.8188} } & 
{\bf \underline{0.7981} } \\ \hline
    \end{tabular}
\end{table}

We built image denoising networks with different activation functions following papers~\cite{zhang2017beyond,zhang2017learning,uchida2018non}.
 An dilated convolution~\cite{DilatedConv2016} and a skip connection~\cite{kim2016accurate} are used as shown in Fig.~\ref{fig:DenoisingNet}.

 The datasets for the training were Yang91~\cite{yang91}, General100~\cite{general100}, and Urban100~\cite{urban100}.
 The patch size of training patch was $64 \times 64$.
 The mini-batch size was 256.
 The input images for the training were generated by adding noise to the ground truth image.
 The standerd deviation was randomly set from 0 to 55.
 The optimizer was Adamax~\cite{Adam2015}.
 We trained 80,000 mini-batches for each activation function.

 We used seven $512 \times 512$ sized images called Lena, Barbara, Boats, F.print, Man, Couple, and Hill for the evaluation~\cite{dabov2006image}.
 The PSNR and SSIM comparisons are summarized in Table~\ref{tab:dconv_psnr} and \ref{tab:dconv_ssim}, respectively, where the bold font represents the best performance and the underline represents higher performance than that of the ReLU.
 Those comparisons demonstrate that the activation functions proposed in the literature~\cite{klambauer2017self,ELU2016,Softplus2010,maas2013rectifier,Swish2017} can not overcome the simple ReLU~\cite{nair2010rectified} in the denoising task in any noise level.
 The performance of the proposed WiG activation function only superiors that of the ReLU in terms of both the PSNR and the SSIM.

\section{Conclusion}
 We have proposed the weighted sigmoid gate unit (WiG) for the activation function of the deep neural networks.
 The proposed WiG consists as the multiplication of the inputs and the weighted sigmoid gate.
 It have been shown that the proposed WiG includes the ReLU and several activation functions proposed in the literature as a special cases.
 We evaluate the WiG with the object recognition task and the image reconstruction task.
 The experimental comparisons that the proposed WiG overcomes existing activation functions including widely used ReLU.

\appendices

\end{document}